\def\year{2019}\relax
\documentclass[letterpaper]{article} 
\usepackage{aaai19}  
\usepackage{times}  
\usepackage{helvet}  
\usepackage{courier}  
\usepackage{url}  
\usepackage{graphicx}  
\usepackage{mathtools}
\usepackage{subcaption}
\usepackage{tikz}
\usetikzlibrary{fit,positioning,matrix,chains,positioning,decorations.pathreplacing,arrows}
\usepackage{amssymb}

\title{Interaction-aware Factorization Machines for Recommender Systems}
\author{Fuxing Hong, Dongbo Huang, Ge Chen \\
Advertising and Marketing Services, Corporate Development Group, Tencent Inc.\\
cstur4@zju.edu.cn, \{andrewhuang,gechen\}@tencent.com \\
}

\begin{document}
\maketitle
\begin{abstract}
Factorization Machine (FM) is a widely used supervised learning approach by effectively modeling of feature interactions. Despite the successful application of FM and its many deep learning variants, treating every feature interaction fairly may degrade the performance. For example, the interactions of a useless feature may introduce noises; the importance of a feature may also differ when interacting with different features. In this work, we propose a novel model named \emph{Interaction-aware Factorization Machine} (IFM) by introducing Interaction-Aware Mechanism (IAM), which comprises the \emph{feature aspect} and the \emph{field aspect}, to learn flexible interactions on two levels. The feature aspect learns feature interaction importance via an attention network while the field aspect learns the feature interaction effect as a parametric similarity of the feature interaction vector and the corresponding field interaction prototype. IFM introduces more structured control and learns feature interaction importance in a stratified manner, which allows for more leverage in tweaking the interactions on both feature-wise and field-wise levels. Besides, we give a more generalized architecture and propose Interaction-aware Neural Network (INN) and DeepIFM to capture higher-order interactions. To further improve both the performance and efficiency of IFM, a sampling scheme is developed to select interactions based on the field aspect importance. The experimental results from two well-known datasets show the superiority of the proposed models over the state-of-the-art methods.

\end{abstract}

\section{Introduction}
\label{sec:introduction}

Learning the effects of feature conjugations, especially degree-2 interactions, is important for prediction accuracy\cite{chang2010training}. For instance, people often download apps for food delivery at meal-time, which suggests that the (order-2) interaction between the app category and the time-stamp is an important signal for prediction\cite{guo2017deepfm}. Applying a linear model on the explicit form of degree-2 mappings
can capture the relationship between features, where feature interactions can be easily understood and domain knowledge can be absorbed. The widely used generalized linear models (e.g., logistic regression) with cross features are effective for learning on a massive scale. Although the feature vector might have billions of dimensions, each instance will typically have only hundreds of non-zero values, and  FTRL\cite{mcmahan2013ad} can save both time and memory when making predictions. However, feature engineering is an important but labor-intensive and time-consuming work, and the ``cold-start" problem may hurt performance, especially in a sparse dataset, where only a few cross features are observed; the parameters for unobserved cross features cannot be estimated. 

To address the generalization issue, factorization machines (FMs)\cite{rendle2010factorization} were proposed, which factorizes coefficients into a product of two latent vectors to utilize collaborative information and demonstrate superior performance to a linear model based on the explicit form of degree-2 mappings. In FM, unseen feature interactions can be learned from other pairs, which is useful, as demonstrated by the effectiveness of latent factor models\cite{chen2014context,hong2016latent}. In fact, by specifying the input feature vector, FM can achieve the same express capacity of many factorization models, such as matrix factorization, the pairwise interaction tensor factorization model\cite{rendle2010pairwise}, 
and SVD++\cite{koren2008factorization}.

Despite the successful application of FM, two-folds significant shortcomings still exist. (1) \emph{Feature aspect.} On one hand, the interactions of a useless feature may introduce noises. On the other hand, treating every feature interaction fairly may degrade the performance. (2) \emph{Field\footnote{A field can be viewed as a class of features. For instance, two features male and female belong to the field gender.} aspect.} A latent factor\footnote{A variable in a latent vector corresponding to an abstract concept.} may also have different importance in feature interactions from different \emph{fields}. 
Assuming that there is a latent factor indicating the quality of a phone, this factor may be more important to the interaction between a phone brand and a location than the interaction between gender and a location. To solve the above problems, we propose a novel model called \emph{Interaction-aware Factorization Machine} (IFM) to learn flexible interaction importance on both \emph{feature aspect} and \emph{field aspect}.

Meanwhile, as a powerful approach to learning feature representation, deep neural networks are becoming increasingly popular and have been employed in predictive models. For example, Wide\&Deep\cite{cheng2016wide} extends generalized linear models with a multi-layer perceptron (MLP) on the concatenation of selected feature embedding vectors to learn more sophisticated feature interactions. However, in the wide part of the Wide\&Deep model, feature engineering is also required and drastically affects the model performance. 

To eliminate feature engineering and capture sophisticated feature interactions, many works\cite{cao2016multi,wang2017deep} are proposed and some of them have fused FM with MLP. FNN\cite{zhang2016deep} initializes parts of the feed-forward neural network with FM pre-trained latent feature vectors, where FM is used as a feature transformation method. PNN\cite{qu2016product} imposes a product layer between the embedding layer and the first hidden layer and uses three different types of product operations to enhance the model capacity. Nevertheless, both FNN and PNN capture only high-order feature interactions and ignore low-order feature interactions. DeepFM\cite{guo2017deepfm} shares the feature embedding between the FM and deep component to make use of both low- and high-order feature interactions; however, simply concatenating\cite{cheng2016wide,guo2017deepfm} or averaging embedding vectors\cite{wang2015learning,chen2017attentive} does not account for any interaction between features. In contrast to that, NFM\cite{he2017neural} uses a bi-interaction operation that models the second-order feature interactions to maintain more feature interaction information. Unfortunately, the pooling operation in NFM may also cause information loss. To address this problem, interaction importance on both \emph{feature aspect} and \emph{field aspect} is encoded to facilitate the MLP to learn feature interactions more accurately.

The main contributions of the paper include the following:
\begin{itemize}
\item To the best of our knowledge, this work represents the first step towards absorbing field information into interaction importance learning.
\item The proposed interaction-aware models can effectively learn interaction importance and require no feature engineering.
\item The proposed IFM provides insight into which feature interactions contribute more to the prediction at the \emph{field} level.
\item A sampling scheme is developed to further improve both the performance and efficiency of IFM.
\item The experimental results on two well-known datasets show the superiority of the proposed interaction-aware models over the state-of-the-art methods.
\end{itemize}

\section{Factorization Machines}\label{sec:fm}

We assume that each instance has attributions $x=\left\{x_{1},x_{2},...,x_{m}\right\} $ from $n$ fields and a target $y$, where $m$ is the number of features and $x_i$ is the real valued feature in the $i$-th category.
Let $V\text{\ensuremath{\in}}\mathbb{R}^{K\times m}$
be the latent matrix, with column vector
$V_{i}$ representing the $K$-dimensional feature-specific latent feature vector of feature $i$. Then pair-wise  enumeration of non-zero features can be defined as
\begin{equation}
\mathcal{X}=\{(i,j)\mid0<i\le m,0<j\le m,j>i,x_i\neq0,x_j\neq0\}.
\end{equation}

\emph{Factorization Machine} (FM)\cite{rendle2010factorization} is a widely used model that captures all interactions between features using the factorized parameters:
\begin{equation}
\overline{y}=w_0 + \sum_{i=1}^m{w_{i}x_{i}}+\underbrace{\sum_{(i,j)\in \mathcal{X}}w_{ij}x_{i}x_{j}}_\text{pair-wise feature interactions} \text{,}\label{eq:fm}
\end{equation}
where $w_0$ is the global bias, and $w_i$ models the strength of the i-th variable. In addition, FM captures pairwise (order-2) feature interactions effectively as $w_{ij}=\langle V_i,V_j\rangle$, where $\langle\cdot{,}\cdot\rangle$ is the inner product of two vectors; therefore, the parameters for unobserved cross features can also be estimated.

\def\layersep{3.5cm}
\begin{figure*}[t]
\centering
\resizebox{2\columnwidth}{!}{
\begin{tikzpicture}[shorten >=1pt,->,draw=black, node distance=\layersep]
    \tikzstyle{every pin edge}=[<-,shorten <=1pt]
    \tikzstyle{neuron}=[draw,circle,minimum size=17pt,inner sep=1pt]
    \tikzstyle{zeroneuron}=[draw,circle,minimum size=17pt,draw,inner sep=1pt]
    \tikzstyle{embeding}=[draw,rectangle,minimum width=60pt,minimum height=17pt,inner sep=1pt]
    \tikzstyle{fieldembeding}=[draw,rectangle,minimum width=60pt,minimum height=17pt,inner sep=1pt]
    \tikzstyle{input neuron}=[neuron, fill=black!25];
    \tikzstyle{output neuron}=[neuron];
    \tikzstyle{embedding neuron}=[embeding];
    \tikzstyle{hidden neuron}=[neuron];
    \tikzstyle{annot} = [text width=5em, text centered]

    \tikzstyle{attention}=[draw,rectangle,minimum width=80pt,minimum height=20pt,inner sep=1pt]
    \tikzstyle{projection}=[draw,rectangle,minimum width=80pt,minimum height=20pt,inner sep=1pt]
    \tikzstyle{attention_val}=[draw,circle,minimum size=17pt,inner sep=1pt]

    \path[yshift=3.2cm]
        node[zeroneuron, pin=left:$x_1$] (I-1) at (0,-1) {0};
    \path[yshift=3.2cm]
        node[input neuron, pin=left:$x_2$] (I-2) at (0,-2) {1};
    \path[yshift=3.2cm]
        node[zeroneuron, pin=left:$x_3$] (I-3) at (0,-3) {0};
    \path[yshift=3.2cm]
        node[input neuron, pin=left:$x_4$] (I-4) at (0,-4) {0.2};
    \path[yshift=3.2cm]
        node[zeroneuron, pin=left:$x_5$] (I-5) at (0,-5) {0};
    \path[yshift=3.2cm]
        node[input neuron, pin=left:$x_6$] (I-6) at (0,-6) {1};

    \path[yshift=3cm]
        node[embedding neuron] (E-2) at (\layersep,-1 cm) {$V_{2}\cdot x_{2}$};
    \path[yshift=3cm]
        node[embedding neuron] (E-4) at (\layersep,-2 cm) {$V_{4}\cdot x_{4}$};
    \path[yshift=3cm]
        node[embedding neuron] (E-6) at (\layersep,-3 cm) {$V_{6}\cdot x_{6}$};

        \node (dots) at (0, -3.5) {$\vdots$};
    \def\pairlayersep{3cm}
    \path[yshift=\pairlayersep]
        node[embedding neuron] (P-2) at (2.3*\layersep,-1 cm) {$(V_2 \odot V_4)x_2x_4$};
    \path[yshift=\pairlayersep]
        node[embedding neuron] (P-4) at (2.3*\layersep,-2 cm) {$(V_2 \odot V_6)x_2x_6$};
    \path[yshift=\pairlayersep]
        node[embedding neuron] (P-6) at (2.3*\layersep,-3 cm) {$(V_4 \odot V_6)x_4x_6$};

    \foreach \source in {2,4,6}
        \path (I-\source.east) edge (E-\source.west);

    \path (E-2.east) edge[right] (P-2.west);
    \path (E-4.east) edge[right] (P-2.west);

    \path[yshift=4cm]
        node[attention] (A) at (4*\layersep,-1 cm) {Attention Net};

    \def\tsep{\pairlayersep}
    \path[yshift=\tsep]
        node[attention_val] (A-2) at (3.5*\layersep,-1 cm) {$T_{24}$};

    \path (E-2.east) edge (P-4.west);
    \path (E-6.east) edge (P-4.west);
    \path[yshift=\tsep]
        node[attention_val] (A-4) at (3.5*\layersep,-2 cm) {$T_{26}$};

    \path (E-4.east) edge (P-6.west);
    \path (E-6.east) edge (P-6.west);
    \path[yshift=\tsep]
        node[attention_val] (A-6) at (3.5*\layersep,-3 cm) {$T_{46}$};

    \foreach \source in {2,4,6} {
        \path (P-\source.east) edge (A.west);
        \path (A.south) edge (A-\source.east);
        \path (P-\source.east) edge (A-\source.west);
    }

    \path[yshift=0cm]
        node[fieldembeding] (U-2) at (\layersep,-1 cm) {$U_{f_2}$};
    \path[yshift=0cm]
        node[fieldembeding] (U-4) at (\layersep,-2 cm) {$U_{f_4}$};
    \path[yshift=0cm]
        node[fieldembeding] (U-6) at (\layersep,-3 cm) {$U_{f_6}$};

    \foreach \source in {2,4,6}
        \path (I-\source.east) edge (U-\source.west);

    \def\fieldpairlayersep{0cm}    
    \path[yshift=\fieldpairlayersep]
        node[fieldembeding] (UU-2) at (2.3*\layersep,-1 cm) {$U_{f_2} \odot U_{f_4}$};
    \path[yshift=\fieldpairlayersep]
        node[fieldembeding] (UU-4) at (2.3*\layersep,-2 cm) {$U_{f_4} \odot U_{f_6}$};
    \path[yshift=\fieldpairlayersep]
        node[fieldembeding] (UU-6) at (2.3*\layersep,-3 cm) {$U_{f_4} \odot U_{f_6}$};

    \path[yshift=\fieldpairlayersep]
        node[projection] (Projection) at (2.8*\layersep,-4 cm) {Projection Matrix};

    \path (U-2.east) edge[right] (UU-2.west);
    \path (U-4.east) edge[right] (UU-2.west);
    \path[yshift=\fieldpairlayersep]
        node[fieldembeding] (F-2) at (3.5*\layersep,-1 cm) {$F_{24}$};

    \path (U-2.east) edge (UU-4.west);
    \path (U-6.east) edge (UU-4.west);
    \path[yshift=\fieldpairlayersep]
        node[fieldembeding] (F-4) at (3.5*\layersep,-2 cm) {$F_{26}$};

    \path (U-4.east) edge (UU-6.west);
    \path (U-6.east) edge (UU-6.west);

    \path[yshift=\fieldpairlayersep]
        node[fieldembeding] (F-6) at (3.5*\layersep,-3 cm) {$F_{46}$};

    \path (Projection.north) edge[right] (F-2.west);
    \path (Projection.north) edge[right] (F-4.west);
    \path (Projection.north) edge[right] (F-6.west);

    \foreach \name / \y in {2,4,6} {
        \path (UU-\name.east) edge[right] (F-\name.west);
        
    }

    \path[yshift=1.5cm]
        node[hidden neuron] (H-2) at (4.5*\layersep,-1 cm) {};
    \path[yshift=1.5cm]
        node[hidden neuron] (H-4) at (4.5*\layersep,-2 cm) {};
    \path[yshift=1.5cm]
        node[hidden neuron] (H-6) at (4.5*\layersep,-3 cm) {};

    \node[output neuron, right of=H-4, node distance=0.8*\layersep] (O) {$\overline{y}$};

    \foreach \source in {2,4,6} {
        \path (A-\source.east) edge[right] (H-\source.west);
        \path (F-\source.east) edge[right] (H-\source.west);
        \path (H-\source.east) edge (O);
    }

    \node[annot,below of=I-6, node distance=2.5cm] (i2) {Input Layer};
    \node[annot,right of=i2, node distance=1*\layersep](e2) {Embedding Layer};
    \node[annot,right of=e2, node distance=1.3*\layersep](p2) {Pair-wise Interaction Layer};
    \node[annot,right of=p2, node distance=1.2*\layersep](inf2) {Inference Layer};
    \node[annot,right of=inf2, node distance=1*\layersep](f2) {Fusion Layer};
    \node[annot,right of=f2, node distance=0.8*\layersep](output) {Output Layer};

\end{tikzpicture}
}
\caption{The neural network architecture of the proposed Interaction-aware Factorization Machine (IFM).}
\label{fig:ifm}
\end{figure*}
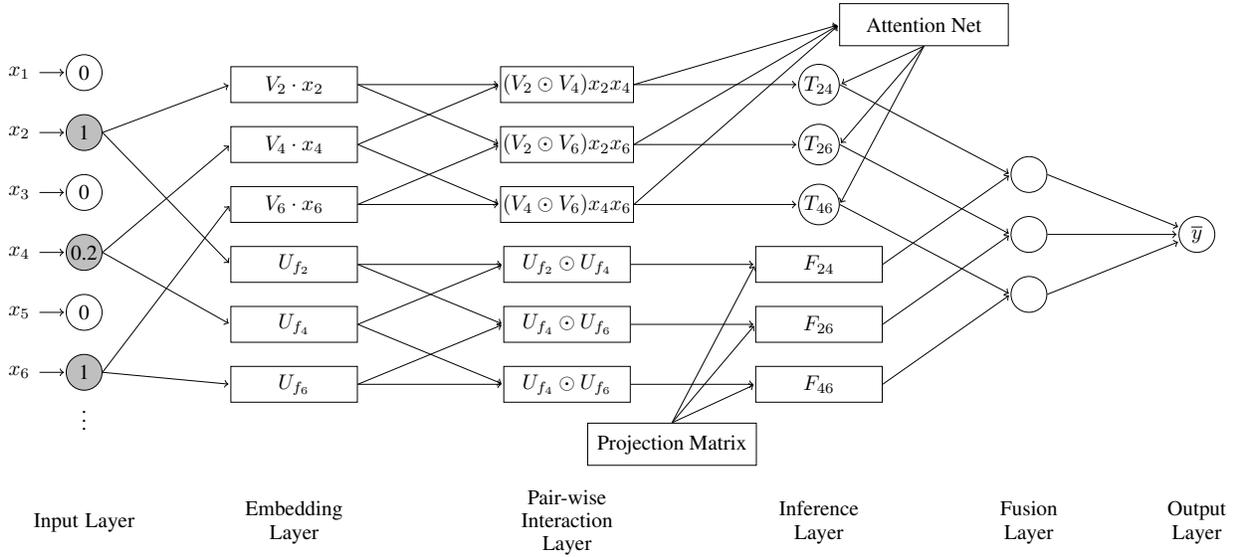
\section{Proposed Approach} \label{sec:proposed}
In this section, we first present the interaction-aware mechanism. Subsequently, we detail the proposed \emph{Interaction-aware Factorization Machine} (IFM). Finally, we propose a generalized interaction-aware model and its neural network specialized versions. 

\subsection{Interaction-Aware Mechanism (IAM)}
The pair-wise feature interaction part of FM can be reformulated as 
\begin{align}
\begin{split}
\sum_{i=1}^m\sum_{j=i+1}^{m} 1 \cdot \langle \mathbf{1}, V_i \odot V_j \rangle x_{i}x_{j}, \\ 
\end{split}
\label{eq:refm}
\end{align}
where $\mathbf{1}$ is a $K$-dimensional vector with all entries one and $\odot$ denotes the Hadamard product. Then we introduce the Interaction-Aware Mechanism (IAM) to discriminate the importance of feature interactions, which simultaneously considers field information as auxiliary information,
\begin{equation}
\sum_{i=1}^m\sum_{j=i+1}^{m}T_{ij}\langle \underbrace{F_{f_i,f_j}}_\text{\clap{field aspect~}}, V_i\odot V_j\rangle x_{i}x_{j} \text{,}
\label{eq:iam}
\end{equation}
where $f_i$ is the field of feature $i$, $F_{f_i,f_j}$ is the $K$-dimensional field-aware factor importance vector of the interaction between feature $i$ and feature $j$ modeling the field aspect; thus, both factors from the same feature interaction and the same factor of interactions from different fields can have significantly different influences on the final prediction. $T_{ij}$ is the corresponding attention score modeling the feature aspect; thus, the importance of feature interactions can be significantly different, which is defined as
\begin{align}
\begin{split}
a_{ij}'&=h^T\mathrm{Relu}(W(V_i \odot V_j)x_i x_j + b), \\
T_{ij}&=\frac{\text{exp}(a_{ij}' / \tau)}{\sum_{(i,j)\in \mathcal{X}}\text{exp}(a_{ij}' / \tau)},
\end{split}
\label{eq:t}
\end{align}
where $K_a$ is the hidden layer size of the attention network, $b\text{\ensuremath{\in}}\mathbb{R}^{K_a}$, $h\text{\ensuremath{\in}}\mathbb{R}^{K_a}$, $W\text{\ensuremath{\in}}\mathbb{R}^{K_a\times K}$, and $\tau$ is a hyperparameter that was originally used to control the randomness of predictions by scaling the logits before applying softmax\cite{hinton2015distilling}. Here we use $\tau$ to control the effectiveness strength of the feature aspect. For a high temperature($\tau \to \infty$), all interactions have nearly the same importance, and the feature aspect has a limited impact on the final prediction. For low temperatures ($\tau \to 0$), the probability of the interaction vector with the highest expected reward tends to 1 and the other interactions are ignored.

The raw presentation of $F$ has $n(n-1)/2 \times K$ parameters, where $n$ is the number of fields, so the space complexity of IAM is quadratic in the field number. We further factorize tensor $F$ using canonical decomposition\cite{kolda2009tensor}:

\begin{equation}
F_{f_i, f_j} = D^T(U_{f_i} \odot U_{f_j}) \text{,}
\label{eq:f}
\end{equation}
where $U\text{\ensuremath{\in}}\mathbb{R}^{ K_F \times n}$ and $D\text{\ensuremath{\in}}\mathbb{R}^{K_F \times K}$, and $K_F$ is the number of latent factors of both $U$ and $D$. Therefore, the space complexity is reduced to $O(nK_F + K_FK)$, which is linear in the field number.

From another perspective, field aspect learns feature interaction effect as a parametric similarity of the feature interaction vector $(V_i\odot V_j)x_ix_j$ and the corresponding field interaction prototype $U_{f_i} \odot U_{f_j}$, which has a bi-linear form\cite{chechik2010large},
\begin{equation}
sim_D(c,e)=c^TDe\text{,}
\end{equation}
with $c=U_{f_i} \odot U_{f_j}$, $e=(V_i\odot V_j)x_ix_j$.

\subsection{Interaction-aware Factorization Machines (IFMs)}

Interaction-aware Factorization Machine (IFM) models feature interaction importance more precisely by introducing IAM. For simplicity, we omit linear terms and the bias term in the remaining parts. Figure~\ref{fig:ifm} shows the neural network architecture of IFM, which comprises 6 layers. In the following, several layers are detailed:
\begin{itemize}
\item \textbf{Embedding layer.} The embedding layer is a fully connected layer that projects each feature to a dense vector representation. IFM employs two embedding matrices $V$ and $U$ for feature embedding and field embedding querying, respectively.

\item \textbf{Pair-wise interaction layer.} The pair-wise interaction layer enumerates interacted latent vectors, each of which is a element-wise product of two embedding vectors from the embedding layer. Let the feature aspect pair-wise interaction set $\mathcal{P}_F$ and the field aspect pair-wise interaction set $\mathcal{P}_I$ be
\begin{equation}
\begin{aligned}
\mathcal{P}_F &= \{(V_i \odot V_j)x_ix_j\mid (i,j)\text{\ensuremath{\in}}\mathcal{X}\}, \\
\mathcal{P}_I &= \{U_{f_i} \odot U_{f_j}\mid (i,j)\text{\ensuremath{\in}}\mathcal{X}\},
\end{aligned}
\label{eq:p}
\end{equation}
then each has no information overlap; the former only depends on the feature embedding matrix $V$, while the latter only comes from the field embedding matrix $U$.
\item \textbf{Inference layer.} The inference layer calculates the \emph{feature aspect} importance and the \emph{field aspect} importance according to Equation~\ref{eq:t} and Equation~\ref{eq:f}, respectively.
\end{itemize}
To summarize, we give the overall formulation of IFM as:
\begin{equation}
\begin{aligned}
\overline{y}= \sum_{i=1}^m\sum_{j=i+1}^{m}T_{ij}(U_{f_i} \odot U_{f_j})^T&D(V_i\odot V_j)x_{i}x_{j} \\
      + &\sum_{i=1}^m{w_{i}x_{i}} + w_0
 \text{.}
\label{eq:ifm}
\end{aligned}
\end{equation}
We also apply $L_2$ regularization on $U$ and $D$ with $\lambda_F$ controling the regularization strength and employ dropout\cite{srivastava2014dropout} on the pair-wise interaction layer to prevent overfitting. Note that $U\text{\ensuremath{\in}}\mathbb{R}^{ K_F \times n}$ and $V\text{\ensuremath{\in}}\mathbb{R}^{K\times m}$ can have different dimensions; each latent vector of $U$ only needs to learn the effect with a specific field, so usually,
\begin{equation}
K_F \ll  K.
\end{equation}

\textbf{Complexity Analysis.} Feature embedding matrix $V$ require $m \times K$ parameters and field-aware factor importance matrix $F$ requires $n \times K_F + K_F \times K$ parameters after applying Equation~\ref{eq:f}. Besides, the parameters of attention network is $K_a \times K + 2K_a$. Thus, the overall space complexity is $O(n K_F+(K_F+m+K_a) K + 2K_a)$, where $K_F,K_a,K$ and $n$ are small compared to $m$, so the space complexity is similar to that of FM, which is $O(m K)$.

The cost of computing $\mathcal{P}_F$ (Equation \ref{eq:p}) and feature aspect importance are $O(|\mathcal{X}| K)$ and $O(|\mathcal{X}| K K_a)$, respectively. For prediction, because the field-aware factor importance matrix $F$ can be pre-calculated by Equation~\ref{eq:f} and the fusion layer only involves the inner product of two vectors, for which the complexity is $O(|\mathcal{X}| K)$, the overall time complexity is $O(|\mathcal{X}| K K_a)$. 

\textbf{Sampling.} We dynamically sample $c$ feature interactions according to the norms of field-aware factor importance vectors ($F_{f_i,f_j}$) and attention scores are only computed for the sampled interactions. The cost of sampling is $O(n^2 K_F K)$ for a mini-batch data and the computation cost of attention scores is $O(c K K_a)$ for every instance. By sampling, the selection frequency for useless interactions is reduced and the overall time complexity is reduced to $O(c K K_a + \frac{n^2}{batchSize} K_FK)$.

\subsection{Generalized Interaction-aware Model (GIM)}
We present a more generalized architecture named Generalized Interaction-aware Model (GIM) in this section and derive its neural network versions to effectively learn higher order interactions. Let \emph{feature aspect} embedding set $\mathcal{F_X}$  and \emph{field aspect} embedding set $\mathcal{I_X}$ be

\begin{align}
\begin{split}
\mathcal{F_X} &= \{T_{ij} V_i \odot V_jx_ix_j\mid (i,j)\text{\ensuremath{\in}}\mathcal{X}\}, \\
\mathcal{I_X} &= \{D^T(U_{f_i} \odot U_{f_j})\mid (i,j)\text{\ensuremath{\in}}\mathcal{X}\},
\end{split}
\label{eq:AB}
\end{align}
Then, the final prediction can be calculated by introducing function $G$ as
\begin{equation}
\overline{y}=G(\mathcal{F_X}, \mathcal{I_X}).
\end{equation}
Let $\mathcal{F_X}_{i,j}$ and $\mathcal{I_X}_{i,j}$ be the element with index $(i,j)$ in $\mathcal{F_X}$ and $\mathcal{I_X}$, respectively. Then IFM can be seen as a special case of GIM using the following, 
\begin{equation}
G_{IFM}(\mathcal{F_X}, \mathcal{I_X}) = \sum\{\mathcal{I_X}_{i,j}^T\mathcal{F_X}_{i,j} \mid (i,j)\text{\ensuremath{\in}}\mathcal{X}\}.
\end{equation}
Besides, $G$ can be a more complex function to capture the non-linear and complex inherent structure of real-world data. Let
\begin{align}
\begin{split}
h^0=&\mathop{concate}\{\mathcal{I_X}_{i,j} \odot \mathcal{F_X}_{i,j} \mid (i,j)\text{\ensuremath{\in}}\mathcal{X}\}, \\
h_{l}=&f_l(Q_lh_{l-1}+z_l), \\
\end{split}
\end{align}
where $n_l$ is the number of nodes in the $l$-th hidden layer; then, $Q_l \text{\ensuremath{\in}}\mathbb{R}^{n_{l}\times n_{l-1}}$, $z_l \text{\ensuremath{\in}}\mathbb{R}^{n_{l}}$ are parameters for the $l$-th hidden layer, $f_l$ is the activation function for the $l$-th hidden layer, and $h_l\text{\ensuremath{\in}}\mathbb{R}^{n_l}$ is the output of the $l$-th hidden layer. Specially, Interaction-aware Neural Network (INN) is defined as
\begin{equation}
G_{INN}(\mathcal{F_X}, \mathcal{I_X}) = h_L,
\label{eq:G_INN}
\end{equation}
where $L$ denotes the number of hidden layers and $f_L$ is the identity function. For hidden layers, we use $\mathrm{Relu}$ as the activation function, which empirically shows good performance.

To learn both high- and low-order feature interactions, the wide component
of DeepFM\cite{guo2017deepfm} is replaced by $G_{IFM}(\mathcal{F_X}, \mathcal{I_X})$ and named as DeepIFM. 

\section{Experimental results}
In this section, we evaluate the performance of the proposed IFM, INN and DeepIFM on two real-world datasets and examine the effect of different parts of IFM. We conduct experiments with the aim of answering the following questions:
\begin{itemize}
\item \textbf{RQ1} How do IFM and INN perform compared to the state-of-the-art methods?
\item \textbf{RQ2} How do the \emph{feature aspect} and the \emph{field aspect} (with sampling) impact the prediction accuracy?
\item \textbf{RQ3} How dose factorization of field-aware factor importance matrix $F$ impact the performance of IFM?
\item \textbf{RQ4} How do the hyper-parameters of IFM impact its performance?

\end{itemize} 
\subsection{Experiment Settings}

\textbf{Datasets and Evaluation.} We evaluate our models on two real-world datasets, MovieLens\footnote{grouplens.org/datasets/movielens/latest}\cite{harper2015movielens} and Frappe\cite{baltrunas2015frappe}, for personalized tag recommendation and context-aware recommendation.
We follow the experimental settings in the previous works\cite{xiaoattentional,he2017neural} and use the optimal parameter settings reported by the authors to have fair comparisons. 
The datasets are divided into a training set (70\%), a probe set (20\%), and a test set (10\%). All models are trained on the training set, and the optimal parameters are obtained on the held-out probe set. The performance is evaluated by the \emph{root mean square error} (RMSE), where a lower score indicates better performance, on the test set with the optimal parameters. Both datasets contain only positive records, so we generate negative samples by randomly pairing two negative samples with each log and converting each log into a feature vector via one-hot encoding. Table~\ref{dataset_desc} shows a description of the datasets after processing, where the sparsity level is the ratio of observed to total features\cite{lee2012comparative}.

\begin{table}[t]
\caption{Dataset Description.} \label{dataset_desc}
\begin{center}
\begin{small} 
\begin{sc} 
\begin{tabular}{lcccr} 
\hline

Data set & MovieLens & Frappe\\ 
\hline 
origin records    & 668,953 & 96,203\\
features & 90,445 & 5,382\\
experimental records & 2,006,859 & 288,609 \\
fields & 3 & 10 \\ 
sparsity level & 0.01\% & 0.19\% \\
\hline \end{tabular} \end{sc} \end{small} \end{center} \end{table} 

\textbf{Baselines.} We compare our models with the following methods:

\begin{itemize}
\item FM\cite{rendle2010factorization}. As described in Equation \ref{eq:fm}. In addition, dropout is employed on the feature interactions 
to further improve its performance.
\item FFM\cite{juan2016field}. Each feature has separate latent vectors to interact with features from different fields. 
\item AFM\cite{xiaoattentional}. AFM learns one coefficient for every feature interaction to enable feature interactions that contribute differently to the prediction. 
\item Neural Factorization Machines (NFMs)\cite{he2017neural}. NFM performs a non-linear transformation on the latent space of the second-order feature interactions. Batch normalization\cite{ioffe2015batch} is also employed 
to address the covariance shift issue.

\item DeepFM\cite{guo2017deepfm}. DeepFM shares the feature embedding between the FM and the deep component.
\end{itemize} 

\textbf{Regularization.} We use $L_2$ regularization, dropout, and early stopping.

\textbf{Hyperparameters.} The model-independent hyperparameters are set to the optimal values reported by the previous works\cite{xiaoattentional,he2017neural}. The embedding size of features is set to 256, and the batch size is set to 4096 and 128 for MovieLens and Frappe, respectively. We also pre-train the feature embeddings
with FM to get better results. For IFM and INN, we set $\tau=10$ and tune the other hyperparameters on the probe set.

\subsection{Model Performance (\textbf{RQ1})}

The performance of different models on the MovieLens dataset and the Frappe dataset is shown in Table~\ref{table:model_performance}, from which the following observations may be made:
\begin{itemize}
\item Learning the importance of different feature interactions improves performance. This observation is derived from the fact that both AFM and the IAM-based models (IFM and INN) perform better than FM does. As the best model, INN outperforms FM by more than 10\% and 7\% on the MovieLens and Frappe datasets, respectively.
\item IFM makes use of field information and can model feature interactions more precisely. To verify the effectiveness of field information, we conduct experiments with FFM and FFM-style AFM, where each feature has separate latent vectors to interact with features from different fields, on the MovieLens dataset. As expected, the utilization of field information brings improvements of approximately 2\% and 3\% with respect to FM and AFM. 
\item INN outperforms IFM by using a more complex function $G$, as described in Equation~\ref{eq:G_INN}, which captures more complex and non-linear relations from IAM encoded vectors. 
\item Overall, our proposed IFM model outperforms the competitors by more than 4.8\% and 1.2\% on the MovieLens and Frappe datasets, respectively. The proposed INN model performs even better, which achieves an improvement of approximately 6\% and 1.5\% on the MovieLens and Frappe datasets, respectively.
\end{itemize}

\begin{table}[t]
\caption{Test RMSE from different models.} \label{table:model_performance}
\begin{center}
\begin{small} 
\begin{sc} 
\scalebox{0.9}{
\begin{tabular}{l|cc|ccr}
\hline
       & Frappe     &  & MovieLens &  &    \\
\hline 
Method & \#Param & RMSE & \#Param   & RMSE \\
\hline
FM     & 1.38M    & 0.3321  & 23.24M    & 0.4671 \\
DeepFM & 1.64M     & 0.3308 & 23.32M    & 0.4662 \\
FFM    & 13.8M    & 0.3304  & 69.55M    & 0.4568 \\
NFM    & 1.45M    &0.3171  & 23.31M     & 0.4549 \\
AFM       & 1.45M     &0.3118    & 23.25M   &0.4430    \\
\textbf{IFM-sampling}       & 1.46M      & \textbf{0.3085}    & -  & -\\
\textbf{IFM}       & 1.46M      & \textbf{0.3080}    & 23.25M  & \textbf{0.4213}\\
\textbf{INN}  &  1.46M   &\textbf{0.3071}   & 23.25M  & \textbf{0.4188}  \\
\hline \end{tabular}} \end{sc} \end{small} \end{center} \end{table}

\subsection{Impact of different aspects and sampling (\textbf{RQ2})} \label{sec:aspect}
\begin{figure}[t!]
\begin{center} 
\begin{subfigure}[MovieLens]{0.47\columnwidth}
{
\includegraphics[width=\columnwidth]{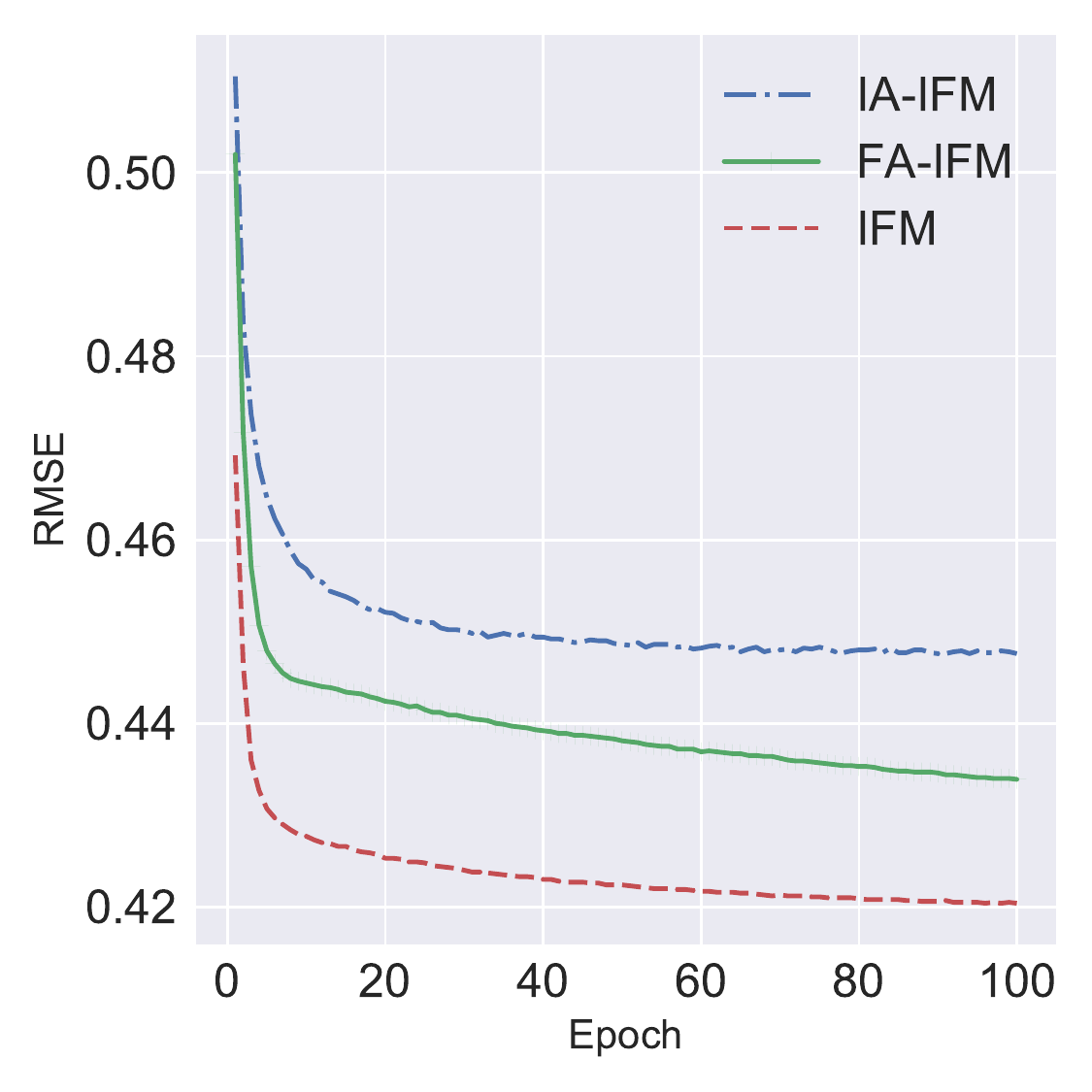}
\caption{MovieLens}
}
\end{subfigure} 
\begin{subfigure}[Frappe]{0.47\columnwidth}
{
\includegraphics[width=\columnwidth]{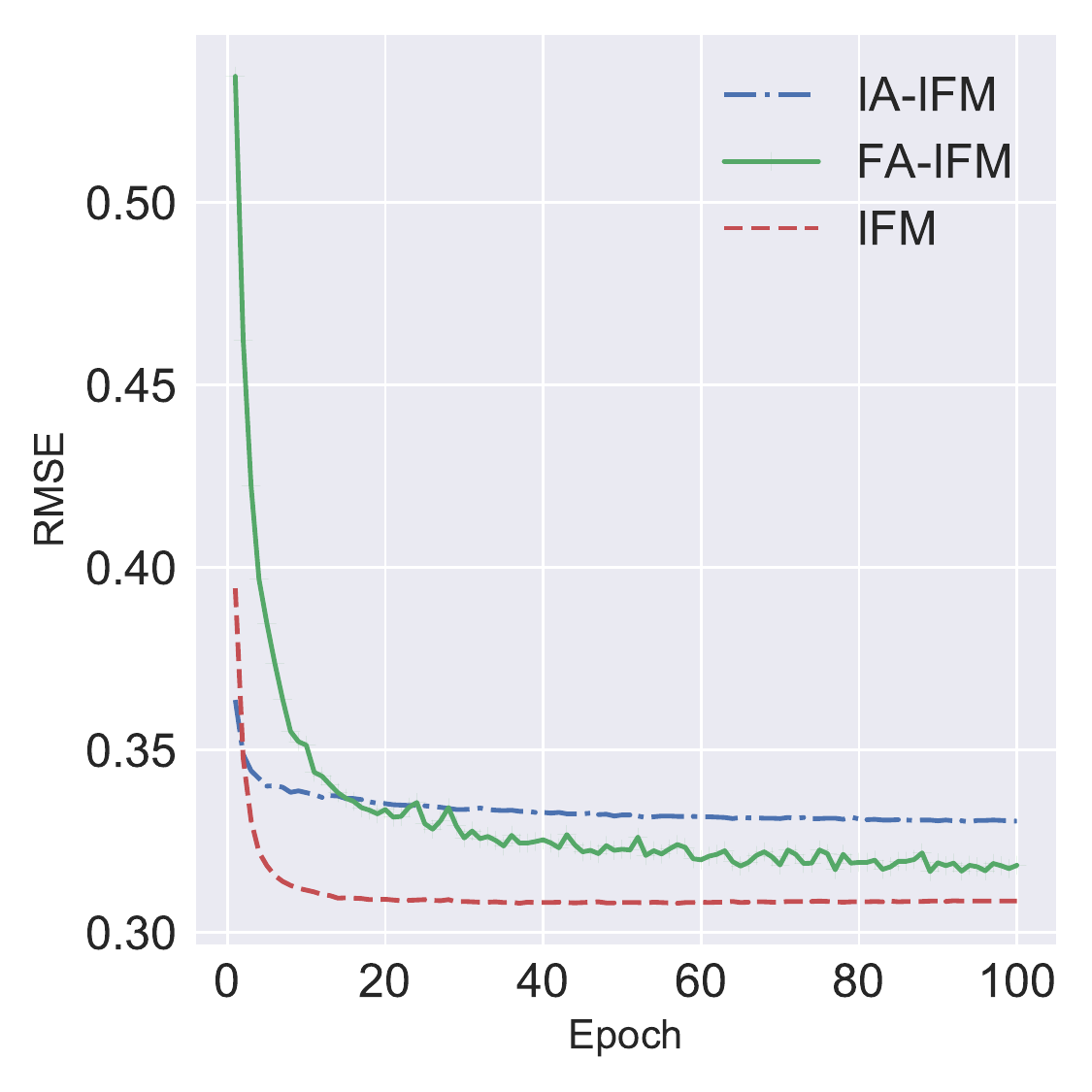}
\caption{Frappe}
} 
\end{subfigure} 
\caption{ Comparison of test RMSE by using only one aspect.} 
\label{fig:aspect} 
\end{center}
\end{figure}

IFM discriminates feature interaction importance on \emph{feature aspect} and \emph{field aspect}. To study how each aspect influences IFM prediction, we keep only one aspect and monitor how IFM performs. As shown in Figure~\ref{fig:aspect}, feature-aspect-only IFM (FA-IFM) performs better than field-aspect-only IFM (IA-IFM) does. We explain this phenomenon by examining the models. The FA-IFM modeling of feature interaction importance is more detailed for each individual interacted vectors; thus, it can make use of the feature interaction information precisely, whereas IA-IFM utilizes only field-level interaction information and lacks the capacity to distinguish feature interactions from the same fields.
Although FA-IFM models feature interactions in a more precise way, IFM still achieves a significant improvement by incorporating field information, which can be seen as auxiliary information, to give more structured control and allow for more leverage when tweaking the interaction between features.

We now focus on analyzing the different role of field aspect in different datasets. 
We calculated the ratio of the improvements of FA-IFM over IA-IFM, which were 9:1 and 1.7:1 on the Frappe and MovieLens datasets, respectively.
It is determined that field information plays a more significant role in the MovieLens dataset. We explain this phenomenon by examining the datasets. As shown in Table~\ref{dataset_desc}, the MovieLens dataset is sparser than the Frappe dataset, where the field information brings more benefit\cite{juan2016field}.

\textbf{Field importance Analysis.} Field aspect not only improves the model performance but also gives the ability to interpret the importance of feature interactions at the field-factor level. Besides, the norm of field aspect importance vector provides insight into interaction importance at the field level. To demonstrate this, we investigate field aspect importance vectors on the MovieLens dataset. As shown in Table~\ref{table:field_analysis}, the movie-tag interaction is the most important while the user-movie interaction has a negligible impact on the prediction because tags link users and items as a bridge\cite{chen2016capturing} and directly modeling semantic correlation between them is less effective.

\textbf{Sampling.} To examine how sampling affects the performance of IFM, an experiment was conducted on Frappe dataset and because there are only three interactions in MovieLens dataset, sampling is meaningless. As shown in Table~\ref{table:model_performance}, IFM with sampling achieves a similar level of performance. 
To verify how sampling performs when the dataset is large, we compare the performance\footnote{Feature interactions from the same field are discarded and the activation of attention network is set to $\mathrm{tanh}$.} on click-through prediction for advertising in \emph{Tencent video}, which has around 10 billion instances. As shown in Table~\ref{fig:large_dataset}, sampling reduce the training time with no significant loss to the performance.

\begin{table}[t]
\caption{The norm of field aspect importance vector of each feature interaction on the MovieLens dataset.} \label{table:field_analysis}

\begin{center}
\begin{small} 
\begin{sc} 
\begin{tabular}{lcccr} 
\hline
 & User-Movie & User-Tag & Movie-Tag \\ 
\hline 
norm & 0.648 & 5.938 & 9.985 \\ 
proportion & 3.9 \% &  35.8 \% &  60.3 \% \\
\hline \end{tabular} \end{sc} \end{small} \end{center} \end{table} 

\begin{table}[t]
\caption{The performance on click-through prediction for advertising in \emph{Tencent video}.}
\label{fig:large_dataset} 
\begin{center}
\begin{small} 
\begin{sc} 
\begin{tabular}{lcccr} 
\hline
 Method &AUC & Time \\ 
\hline 
DeepIFM  &  0.8436 & 16hrs, 18mins \\ 
DeepIFM-Sampling(10\%) &  0.8420  &  3hrs, 49mins  \\
\hline \end{tabular} \end{sc} \end{small} \end{center} \end{table}

\subsection{Impact of factorization (\textbf{RQ3})}
\begin{figure}[t!]
\begin{center} 
\begin{subfigure}[MovieLens]{0.47\columnwidth}
{
\includegraphics[width=\columnwidth]{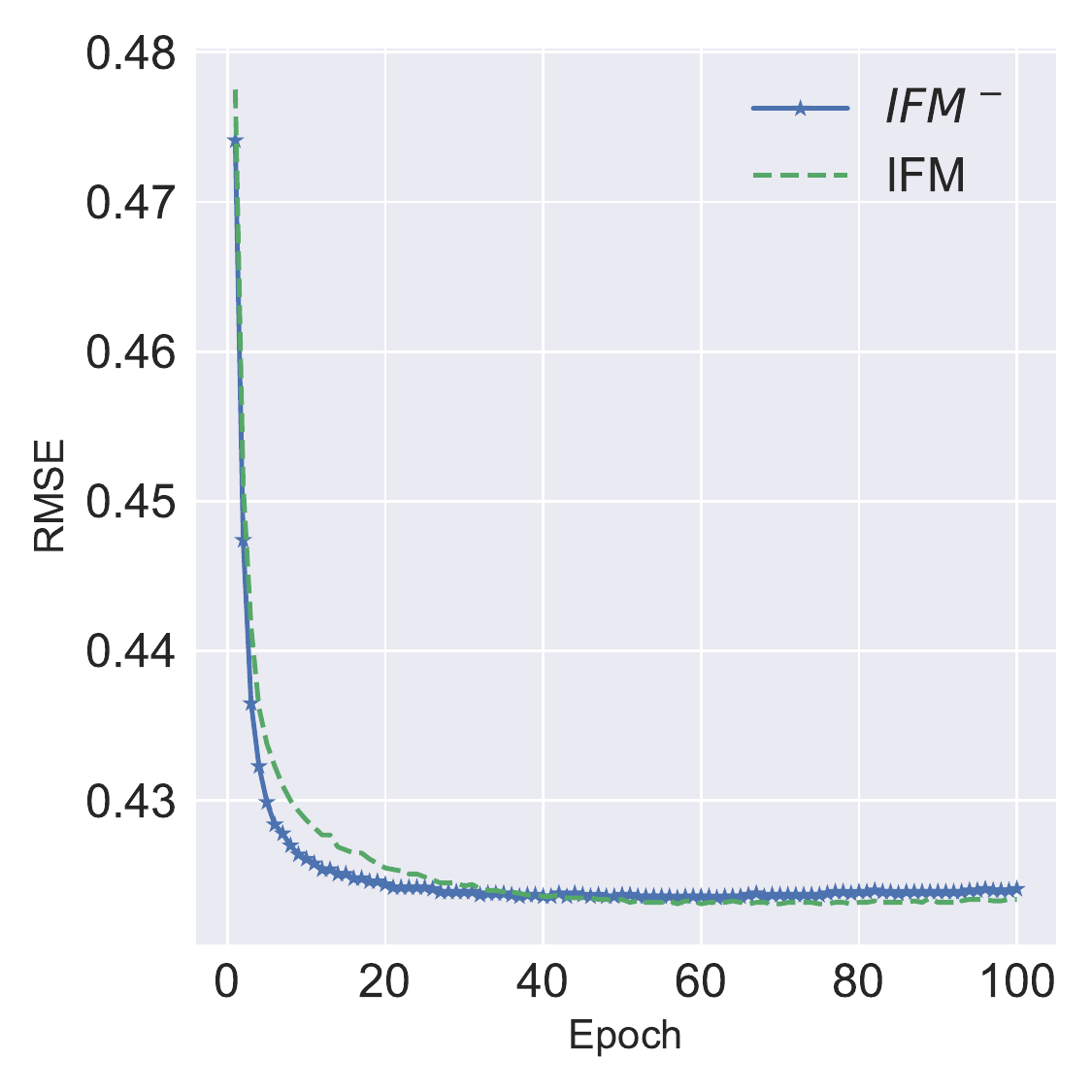}
\caption{MovieLens}
} 
\end{subfigure}
\begin{subfigure}[Frappe]{0.47\columnwidth}
{
\includegraphics[width=\columnwidth]{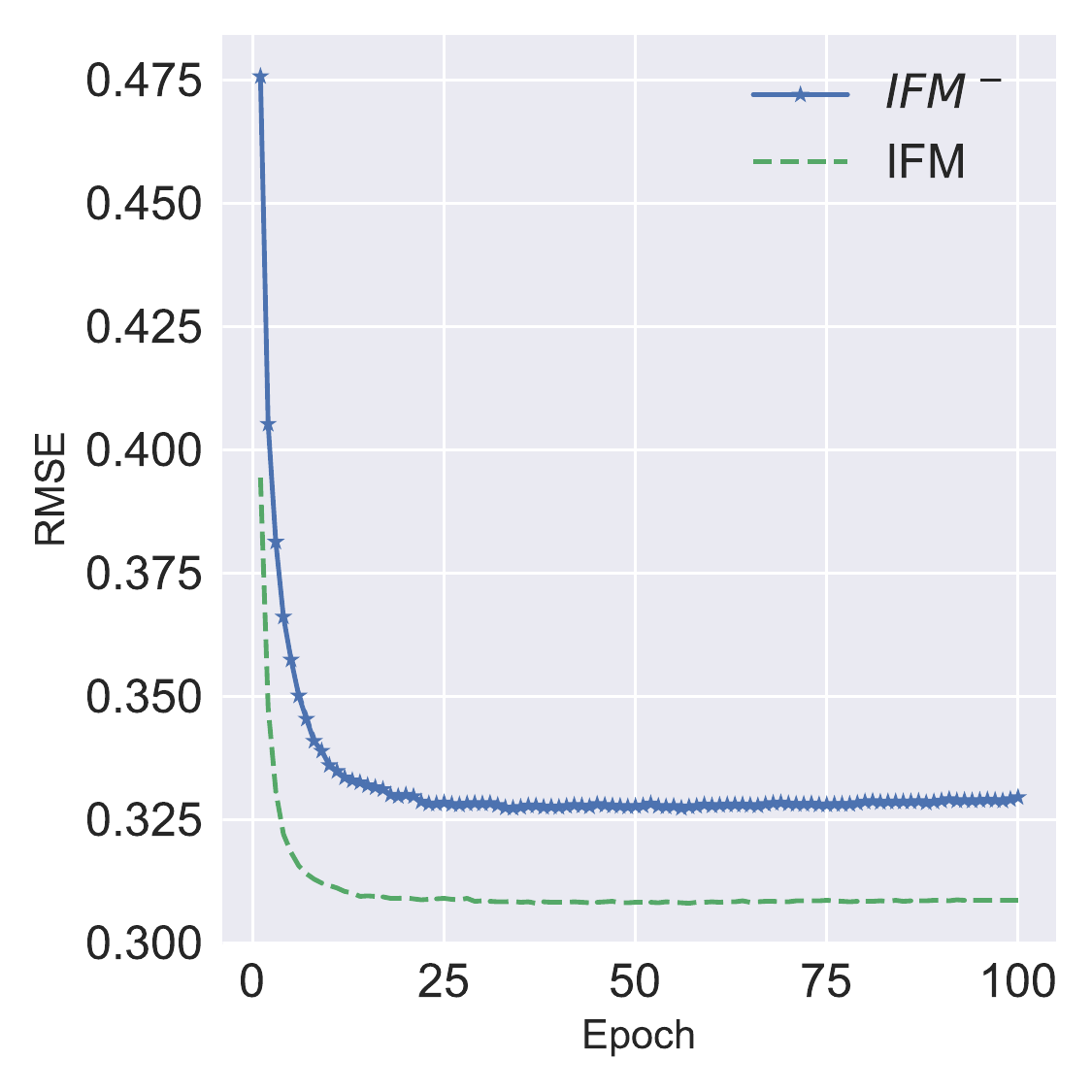}
\caption{Frappe}
} 
\end{subfigure}
\caption{ Performance comparison on the test set \emph{w.r.t.} IFM and the non-factorization version IFM-.}
\label{fig:factorization_nonfactorization} 
\end{center}
\end{figure}

As described in Equation~\ref{eq:f}, IAM factorizes field-aware factor importance matrix $F\text{\ensuremath{\in}}\mathbb{R}^{ n(n-1)/2 \times K}$ to get a more compact representation. We conduct experiments with both the factorized version and the non-factorized version (indicated as IFM$^-$)
 to determine how factorization affects the performance. As shown in Figure~\ref{fig:factorization_nonfactorization}, factorization can speed up the convergence of both datasets. However, it also has a significantly different impact on the performance of the two datasets. For the MovieLens dataset, both versions achieve similar levels of performance but IFM outperforms IFM$^-$ by a large margin on the Frappe dataset, where the performance of IFM$^-$ is degraded from epoch 50 because of an overfitting issue\footnote{Early stopping is disabled in this experiment.}. We explain this phenomenon by comparing the number of entries of field-aware factor importance matrix $F$. For the Frappe dataset, IFM$^-$ and IFM have 11,520 and 6,370 entries with the optimal settings with $K=256$ and $K_F=26$, respectively. That is, after factorization, we can reduce more than 44\% of the parameters, thereby significantly reducing the model complexity. In contrast to that, the MovieLens dataset contains only three interactions, where the effect of factorization is negligible and IFM$^-$ performs slightly better than IFM does although the gap is negligible, i.e., around 0.1\%.

\subsection{Effect of Hyper-parameters (\textbf{RQ4})}

\begin{figure}[t!]
\begin{center} 
\begin{subfigure}[MovieLens]{0.47\columnwidth}
{
\includegraphics[width=\columnwidth]{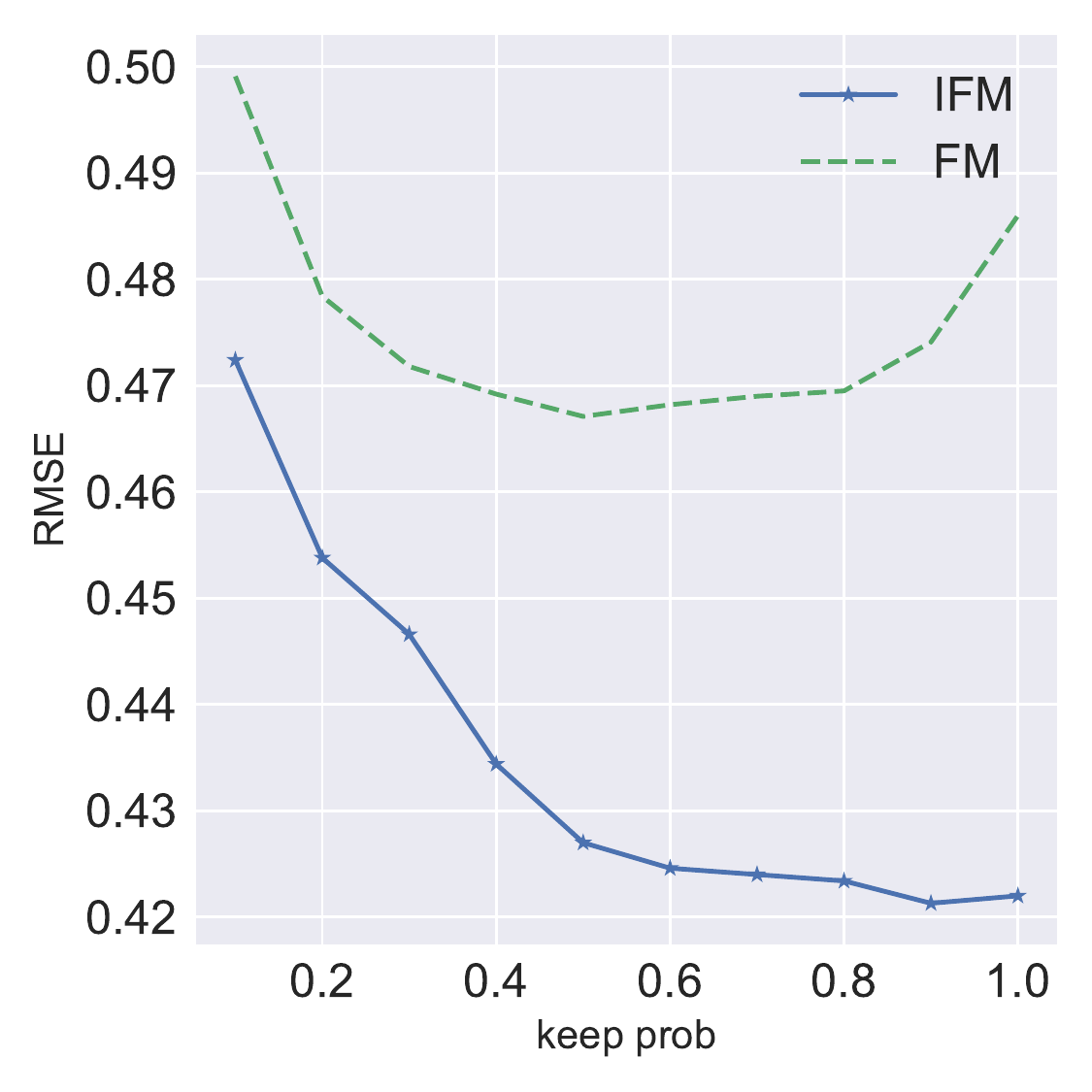}
\caption{MovieLens}
} 
\end{subfigure}
\begin{subfigure}[Frappe]{0.47\columnwidth}
{
\includegraphics[width=\columnwidth]{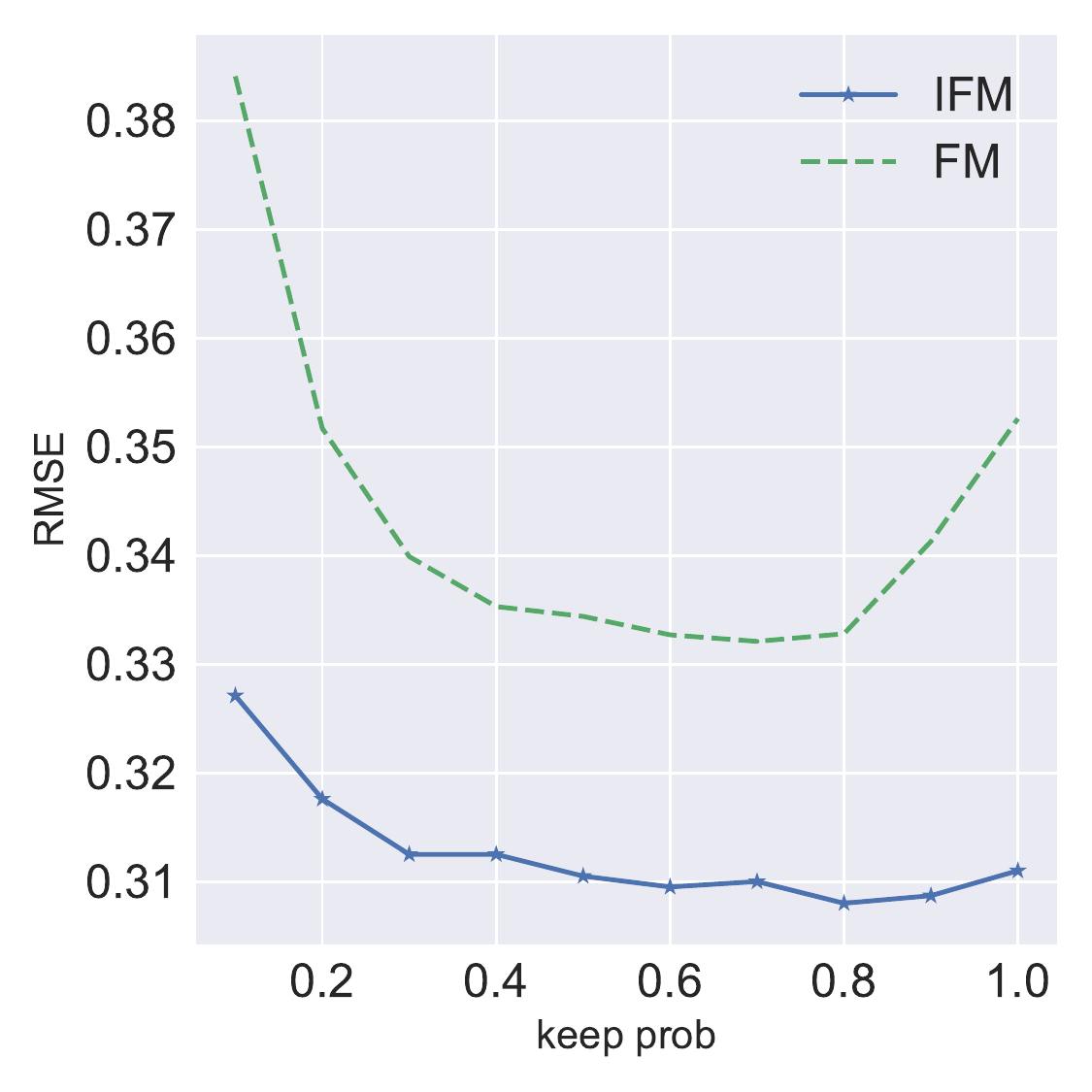}
\caption{Frappe}
} 
\end{subfigure}
\caption{ Comparison of test RMSE by varying keep probabilities.} 
\label{fig:dropout}  
\end{center}
\end{figure}

\begin{figure}[t!]
\begin{center} 
\begin{subfigure}[MovieLens]{0.47\columnwidth}
{
\includegraphics[width=\columnwidth]{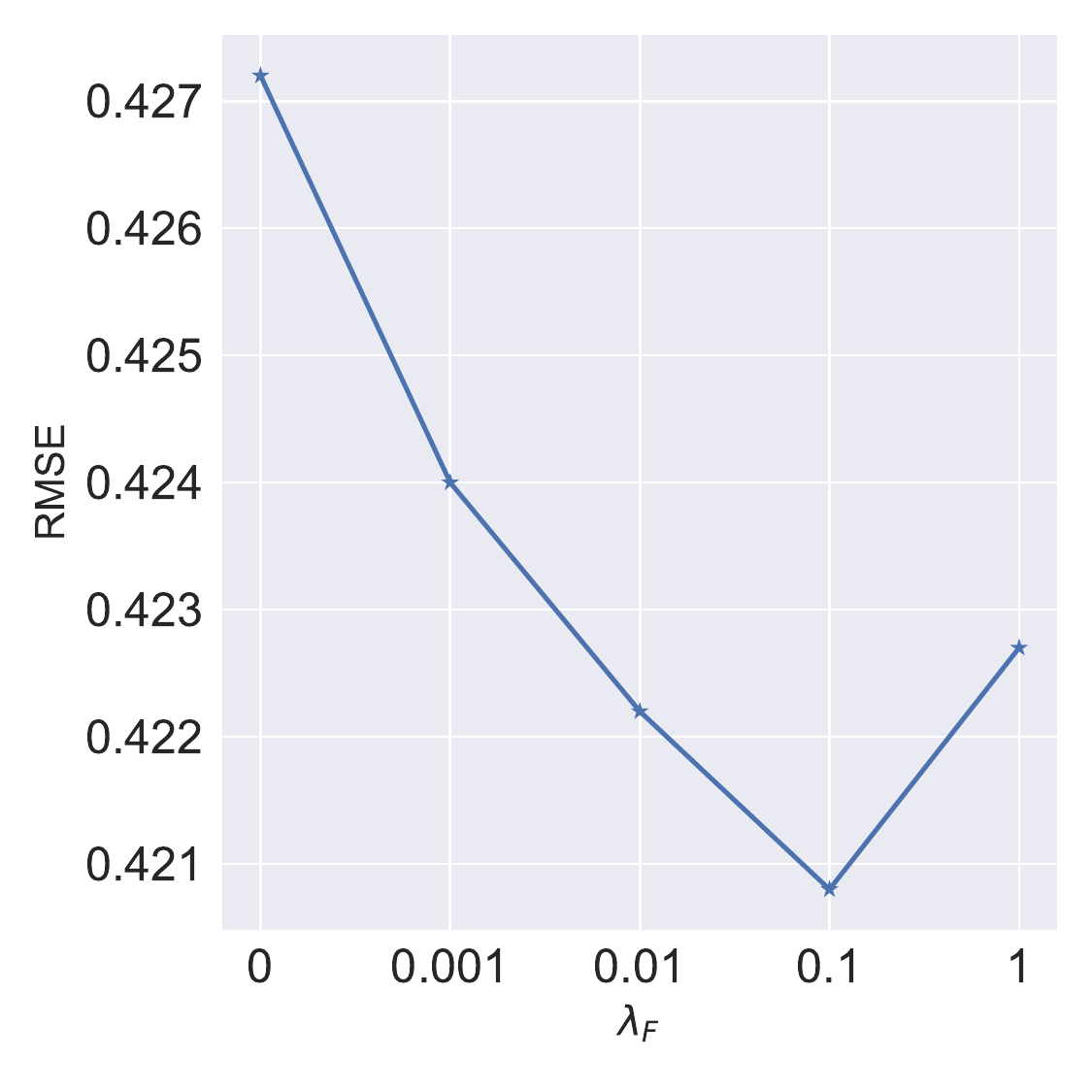}
\caption{MovieLens}
} 
\end{subfigure} 
\begin{subfigure}[Frappe]{0.47\columnwidth}
{
\includegraphics[width=\columnwidth]{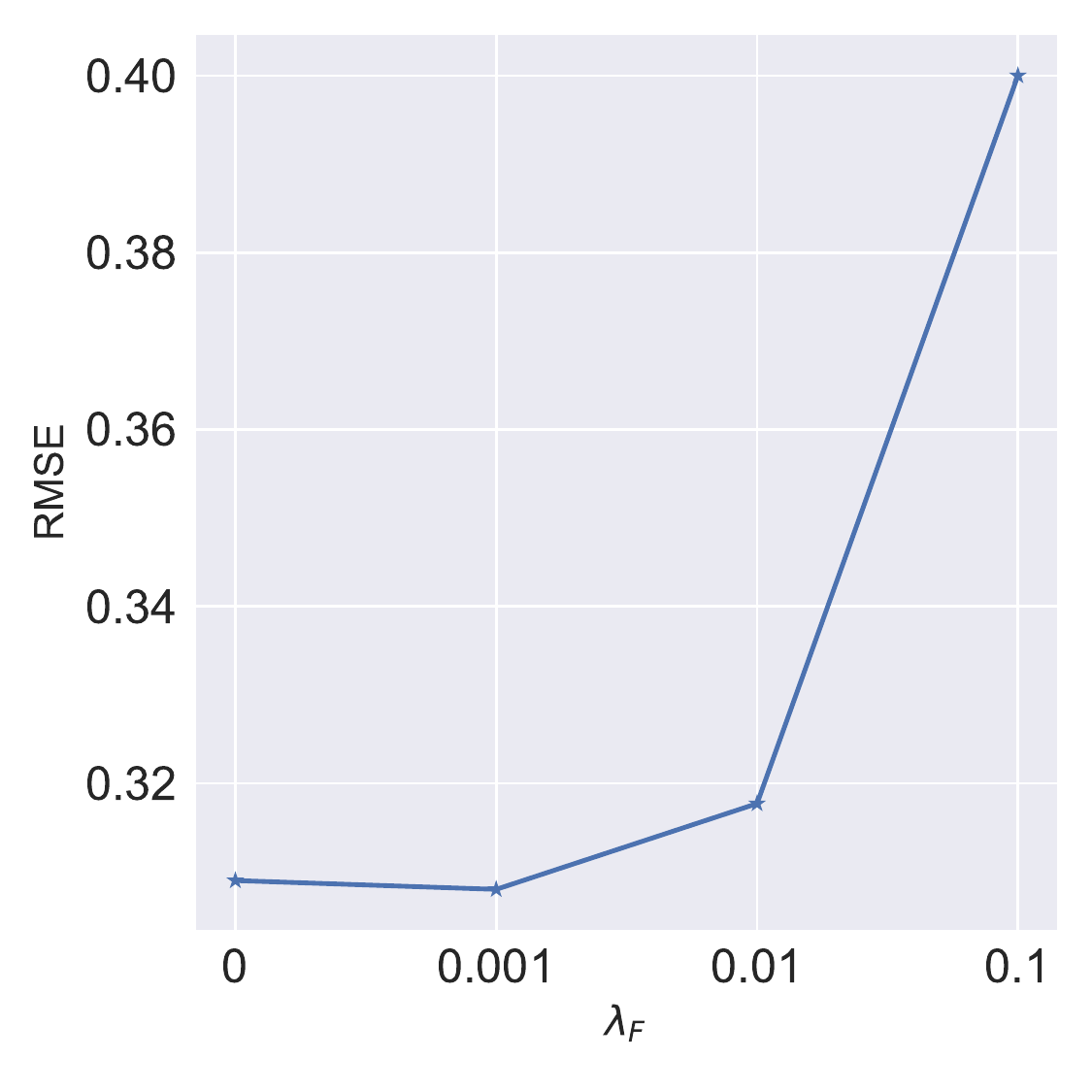}
\caption{Frappe}
}
\end{subfigure} 
\caption{ Comparison of test RMSE by varying $\lambda_F$.}
\label{fig:regularization} 
\end{center}
\end{figure} 

\textbf{Dropout.} Dropout can be seen as a model averaging approach to reduce overfitting by preventing complex co-adaptations on training data. We apply dropout to FM on feature interaction vectors and obtain better performance as a benchmark. As shown in Figure~\ref{fig:dropout}, we set the keep probability from 0.1 to 1.0 with increments of 0.1 and it significantly affects the performance of both FM and IFM. When the keep probability tends to zero, the performance of both models is poor due to the underfitting issue. When the keep probability tends to 1, i.e., no dropout is employed, both models also cannot achieve the best performance. Both IFM and FM achieve the best performance when the keep probability is properly set due to the extreme bagging effect. For nearly all keep probabilities, IFM outperforms FM, which shows the effectiveness of IAM.

\textbf{$L_2$ regularization.} Figure~\ref{fig:regularization} shows how IFM performs when the $L_2$ regularization hyperparameter $\lambda_F$ varies while keeping the dropout ratio constant (optimal value from the validation dataset). IFM performs better when $L_2$ regularization is applied and it achieves an improvement of approximately 1.4\% in the MovieLens dataset. We explain this phenomenon as the following. Using dropout on the pair-wise interaction layer only prevent overfitting for the feature aspect and $\lambda_F$ controls the regularization strength of factorization parameters for the field aspect importance learning. 

\begin{figure}[t!]
\begin{center} 
\begin{subfigure}[MovieLens]{0.47\columnwidth}
{
\includegraphics[width=\columnwidth]{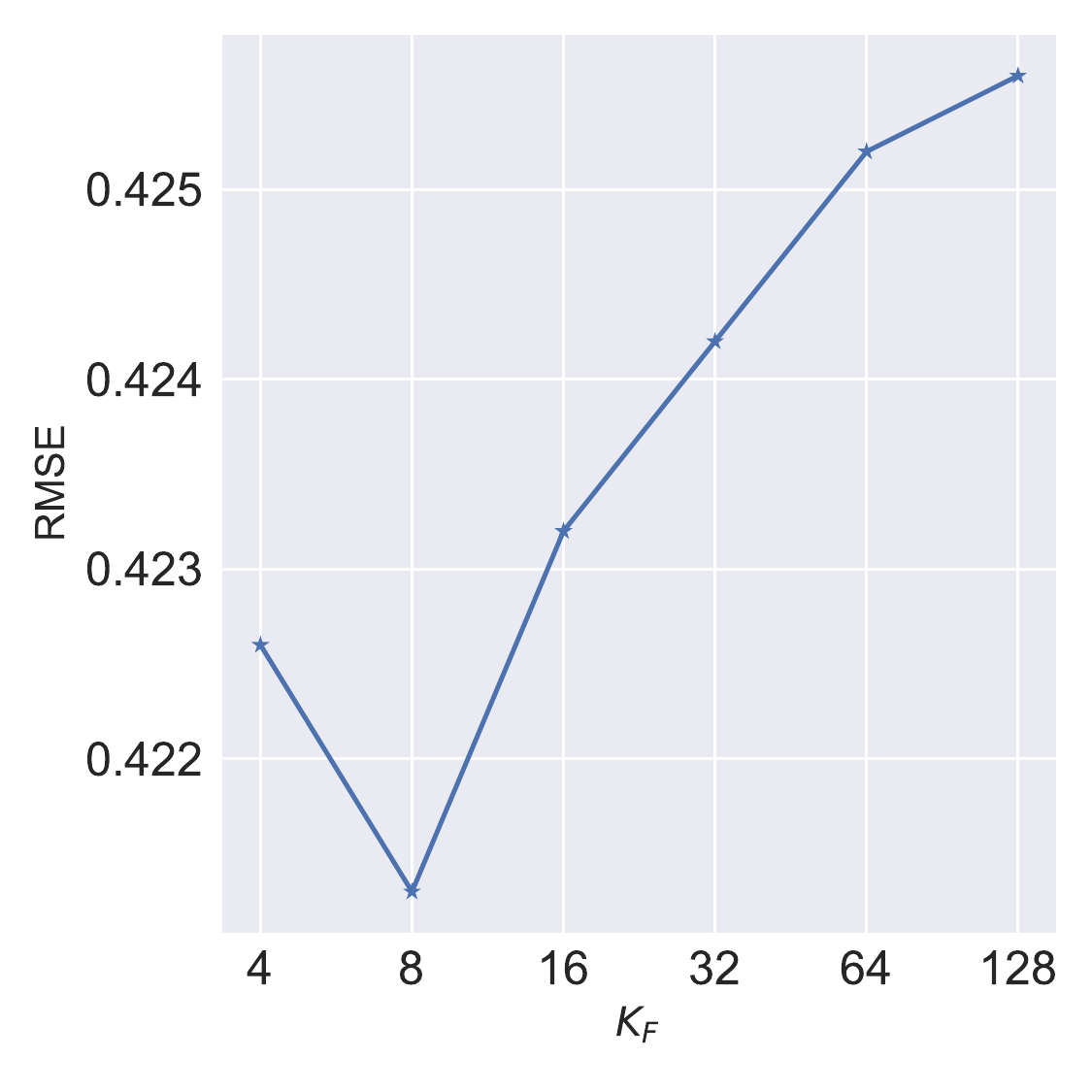}
\caption{MovieLens}
}
\end{subfigure}  
\begin{subfigure}[Frappe]{0.47\columnwidth}
{
\includegraphics[width=\columnwidth]{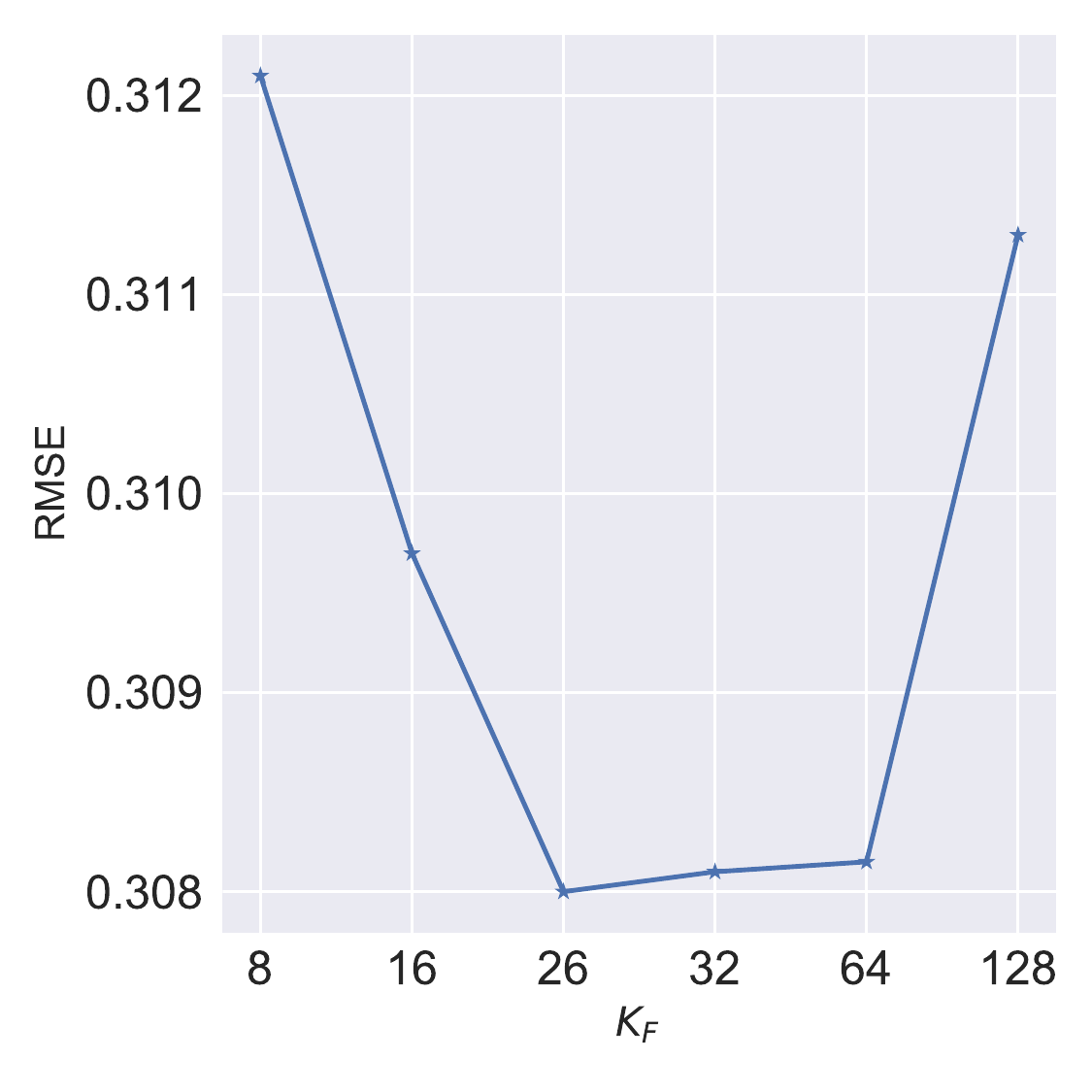}
\caption{Frappe}
} 
\end{subfigure} 
\caption{Comparison of test RMSE by varying $K_F$.} 
\label{fig:kf_hidden} 
\end{center}
\end{figure}

\textbf{The number of hidden factors $K_F$.} Figure~\ref{fig:kf_hidden} shows how IFM performs when the number of hidden factors $K_F$ varies. 
IFM cannot effectively capture the field-aware factor importance when $K_F$ is small and it also can not achieve the best performance when $K_F$ is large due to the overfitting issue. An interesting phenomenon is that the best $K_F$ for the MovieLens dataset is much smaller than that for the Frappe dataset. We explain this phenomenon by looking into the datasets. Because the number of fields $n$ is 10 for the Frappe dataset, the field-aware factor importance matrix captures the importance of factors from 45 interacted vectors. While the MovieLens dataset contains only 3 interactions and the field-aware factor importance matrix keeps much less information. 

\section{Related work} \label{sec:related}
In the introduction section, factorization machine and its many neural network variants are already mentioned, thus we do not discuss them here. In what follows, we briefly recapitulate the two most related models, i.e., AFM\cite{xiaoattentional} and FFM\cite{juan2016field}.

AFM learns one coefficient for every feature interaction to enable feature interactions that contribute differently to the final prediction and the importance of a feature interaction is automatically learned from data without any human domain knowledge.
However, the pooling layer of AFM lacks the capacity of discriminating factor importance in feature interactions from different fields. In contrast, IFM models feature interaction importance at interaction-factor level; thus, the same factor in different interactions can have significantly different influences on the final prediction.

In FMs, every feature has only one latent vector to learn the latent effect with any other features. FFM utilizes field information as auxiliary information to improve model performance and introduces more structured control. In FFM, each feature has separate latent vectors to interact with features from different fields, thus the effect of a feature can differ when interacting with features from different fields. 
However, modeling feature interactions without discriminating importance is unreasonable. IFM learns flexible interaction importance and outperforms FFM by more than 6\% and 7\% on the Frappe and MovieLens datasets, respectively. Moreover, FFM requires $O(m n K)$ parameters, while the space complexity of IFM is $O(m K)$.

\section{Conclusion and Future Directions}

In this paper, we proposed a generalized interaction-aware model and its specialized versions to improve the representation ability of FM. They gain performance improvement based on the following advantages. (1) All models can effectively learn both the feature aspect and the field aspect interaction importance. (2) All models can utilize field information that is usually ignored but useful. (3) All models apply factorization in a stratified manner. (4) INN and DeepIFM can learn jointly with deep representations to capture the non-linear and complex inherent structure of real-world data.

The experimental results on two well-known datasets show the superiority of the proposed models over the state-of-the-art methods.
To the best of our knowledge, this work represents the first step towards absorbing field information into feature interaction importance learning.

In the future, we would like to generalize the field-aware importance matrix to a more flexible structure by applying neural architecture search\cite{liu2017progressive}.

\bibliographystyle{aaai}

\end{document}